\documentclass{article}
\usepackage{cite}

\usepackage{spconf,amsmath,epsfig}
\usepackage{amsfonts}
\usepackage{amssymb}
\usepackage{bbding}
\usepackage{pifont}
\usepackage{hyperref}
\usepackage{titlesec}
\usepackage{hhline}
\usepackage{booktabs}
\usepackage{caption}
\usepackage{multirow}
\usepackage{hyperref} 
\usepackage{utfsym}
\let\OLDthebibliography\thebibliography
\renewcommand\thebibliography[1]{
  \OLDthebibliography{#1}
  \setlength{\parskip}{0pt}
  \setlength{\itemsep}{0pt plus 0.3ex}
}

\title{3DMIT: 3D Multi-modal Instruction Tuning for Scene Understanding}
\name{Zeju Li\textsuperscript{1,2}, Chao Zhang\thanks{* Corresponding author.}\textsuperscript{2}*, Xiaoyan Wang\textsuperscript{2}, Ruilong Ren\textsuperscript{3}, Yifan Xu\textsuperscript{4}, Ruifei Ma\textsuperscript{5}, Xiangde Liu\textsuperscript{2}}

\address{\textsuperscript{1}Beijing University of Posts and Telecommunications, \textsuperscript{2} Beijing Digital Native Digital City Research \\ Center, \textsuperscript{3}Peking University, \textsuperscript{4}Beihang University, \textsuperscript{5}Beijing University of Science and Technology
\\
\textit{\textsuperscript{1}lizeju0727@gmail.com, \textsuperscript{2}zhangchao@bdnrc.org.cn}}

\begin{document}\sloppy

\def\x{{\mathbf x}}
\def\L{{\cal L}}


\maketitle

\begin{abstract}
The remarkable potential of multi-modal large language models (MLLMs) in comprehending both vision and language information has been widely acknowledged. However, the scarcity of 3D scenes-language pairs in comparison to their 2D counterparts, coupled with the inadequacy of existing approaches in understanding of 3D scenes by LLMs, poses a significant challenge. In response, we collect and construct an extensive dataset comprising 75K instruction-response pairs tailored for 3D scenes. This dataset addresses tasks related to 3D VQA, 3D grounding, and 3D conversation. To further enhance the integration of 3D spatial information into LLMs, we introduce a novel and efficient prompt tuning paradigm, 3DMIT. This paradigm eliminates the alignment stage between 3D scenes and language and extends the instruction prompt with the 3D modality information including the entire scene and segmented objects. We evaluate the effectiveness of our method across diverse tasks in the 3D scene domain and find that our approach serves as a strategic means to enrich LLMs' comprehension of the 3D world. Our code is available at \url{https://github.com/staymylove/3DMIT}.
\end{abstract}
\section{Introduction}
\label{sec:intro}


 In recent years, Large Language Models (LLMs) such as LLaMA~~\cite{touvron2023LLaMA}, Vicuna~~\cite{chiang2023Vicuna} and MiniGPT~\cite{zhu2023minigpt} have emerged as powerful tools in natural language processing, demonstrating exceptional capabilities in tasks such as language understanding, generation, and translation. 
LLMs operate on the principle of pre-training followed by fine-tuning on specific tasks. 
This pre-training phase involves exposure to vast amounts of diverse text data, allowing the model to learn the intricacies of language. 
Fine-tuning can tailors the model's knowledge to specific domains or tasks, enhancing its performance in targeted applications. 

In response to the remarkable capabilities demonstrated by LLMs, researchers have extended their scope to develop Multi-modal Language Models (MLLMs). These MLLMs capitalize on the prowess of existing LLMs to address multi-modal applications, which involve the fusion of information
from diverse sources, such as images and 3D point clouds. 
This integration of vision and language not only expands the utility of LLMs but also elevates them into potent tools for a range of tasks, covering visual captioning, visual question answering (VQA), and visual grounding.
In recent studies ~\cite{hong20233d,yin2023lamm,wang2023chat}, LLMs have 
been employed to comprehend the real 3D world. By leveraging 3D scene features and language prompts, researchers have successfully trained 3D multi-modal Language Models (3D-LLMs) for scene understanding. 
Traditional methods which excel in specific tasks, but these models often lack adaptability to other tasks.
In contrast, 3D-LLMs utilize an unified architecture to perform diverse tasks, such as 3D VQA and 3D grounding, 
owing to the reasoning and understanding capabilities inherent in LLMs. The development of 3D-LLMs thus holds great promise.
However, despite the optimistic outlook for 3D-LLMs, there are three challenges: 
1) there exists a scarcity of high-quality 3D scene-language data when compared to the abundance of textual data associated with image and 3D object pairs. 
2) the stage of alignment language with the features of 3D objects and scenes is a time-consuming and laborious process. It is inefficient to train 3D-LLMs in multiple stages. 
3) if training with object-centric 3D instruction and only objects or only the global scene, the 3D-LLMs may not simultaneously grasp global scene information and fine-grained object information.

In this paper, we construct a comprehensive 3D scene-language instruction dataset designed for multi-task applications and we propose 3DMIT, 
an efficient 3D multi-modal instructions tuning method to train LLMs~\cite{chiang2023Vicuna} and MLLMs~\cite{liu2023improved} for multi-task scene understanding by leveraging our presented dataset.
Our instruction dataset builds upon existing datasets, including Scannet~\cite{dai2017scannet} and ScanRefer~\cite{chen2020scanrefer}.
We utilize the original data and employ the GPT-API to generate high-quality instructions and responses for both object-centric and global scene-related tasks. 
Our 3D instruction dataset encompasses 75K 3D scene-language pairs derived from 3D scene dataset: Scannet. Tasks addressed in the dataset include 3D VQA, 3D Captioning, 3D Grounding, and 3D Conversations.

To capture the visual information of the scene, we initially employ a pre-trained scene encoder~\cite{huang2022frozen} $E_S$ to extract features $f_S$ from the point cloud of the scene. While this provides global scene features, there is a potential loss of fine-grained object visual information.
To address this limitation, we subsequently segment the scene and extract the point cloud of each object.
Utilizing a pre-trained 3D encoder~\cite{xue2023Ulip,zhou2023Uni3D} $E_O$ , we extract features $f_O$ of objects within the scene and fuse with the attributes of the objects, complementing the global scene features. 
Notably, different from Chat-3D~\cite{wang2023chat} and 3D-LLM~\cite{hong20233d}, we skip the stage that employs visual features aligned with text embeddings, and directly concatenate visual features and embeddings of text prompts as our 3D multi-modal prompts. 
With instructions of each task, we can fine-tune 3D multi-modal LLMs efficiently.

We also evaluate our method on traditional 3D-language downstream datasets, such as ScanQA~\cite{azuma2022scanqa}, ScanRefer~\cite{chen2020scanrefer} and 3D Multi-choice~\cite{yin2023lamm}. 
Our model outperforms previous 3D-LLMs baselines along with more efficiency.

Our contributions can be summarized as follows:
\begin{itemize}
  \vspace{-5pt}
  \item We construct a comprehensive 3D scene-language instruction dataset with GPT4-API,   
  which encompasses 75K 3D scene-language pairs for 3D VQA, 3D Scene Captioning, 3D Visual Grounding, and 3D Conversation task.
  \vspace{-5pt}
  \item We propose 3DMIT, an efficient 3D multi-modal instructions tuning method to train LLMs. 
   Distinguishing itself from other 3D-LLMs, 3DMIT eliminates the alignment stage between 3D scenes and language, prioritizing efficiency.
  \vspace{-5pt}
  \item We evaluate our method on traditional 3D-language downstream tasks, ScanQA for VQA, ScanRefer for 3D grounding and 3D Multi-choice. 
  Our method outperforms our baselines along with more efficiency.
\end{itemize}

\section{Related work}
\subsection{3D Scene Understanding}
The 3D scene understanding is similar to existing visual question
answering. 
This concept is termed as "3D-language scene understanding" and contains various downstream tasks: 
3D Question Answering~\cite{azuma2022scanqa}, requires models to answer questions based on 3D scenes; 
3D Visual Grounding~\cite{chen2020scanrefer} employs models to localize target objects in 3D scenes according to language queries. 
While traditional methods which excel in specific tasks, but these models often lack adaptability to other tasks.
In contrast, 3D Multi-modal Large language models (3D-LLMs) can utilize an unified architecture to perform 3D scene understanding covering 3D VQA and 3D grounding, 
owing to the reasoning and understanding capabilities inherent in LLMs.

\subsection{3D Multi-modal Large Language Models}
Inspired by remarkable capabilities of LLMs, researchers develop Multi-modal LLMs (MLLMs) by leveraging these models for multi-modal tasks. 
3D-LLM~\cite{hong20233d} can take 3D points with features and language prompts as input, and perform a variety of 3D-related tasks. 
ImageBind-LLM~\cite{han2023imagebind} align 3D point clouds with other modalities within a joint embedding space.
Chat-3D~\cite{wang2023chat} introduce three-stage training scheme to enable the model to transition from learning individual object attributes.
However, it is inefficient to train 3D-LLMs with multiple stages and 3D-LLMs may not simultaneously grasp global scene information and fine-grained object information. 
Our 3DMIT removes the alignment stage of 3D and text, which can make training 3D-LLMs is efficient and effective, with better integration of vision-text information.

\section{Dataset}
Due to the scarcity of high-quality 3D scene-language data, we collect and construct a comprehensive 3D scene-language instruction dataset with GPT4-API. The dataset covers 1514 3D scenes of Scannet 
and encompasses 75K 3D scene-language pairs for 3D VQA, 3D Scene Captioning, 3D Grounding, and 3D Conversation task. Table~\ref{tab:data} presents a detailed breakdown of the dataset, delineating the quantity of data for each respective task.
\begin{table}[h]
\center
\caption{The dataset we bulid to train our LLM} 
\vspace{-6pt}
\label{tab:data}
\scalebox{0.90}{
\begin{tabular}{c|c}
  \toprule
   Tasks & ScanNet\\
  \midrule
 VQA  & 25563   \\
 Scene-level Captioning  & 562  \\
 Conversations  & -  \\
 Grounding  & 36665 \\
 Multiple-choice  & 11895 \\
 All& 75K \\
  \bottomrule
\end{tabular}
}
\vspace{-15pt}
\end{table}



\subsection{3D VQA, Captioning and Conversations}
Based on ScanQA~\cite{azuma2022scanqa} and ScanRefer
~\cite{chen2020scanrefer}, we collect and construct $Instruction-Answer$ pairs with GPT-API. We utilize the caption of each object and generate corresponding questions. 
We use GPT to generate choices for multiple-choice question answering problem on 3D scene dataset.
Furthermore, sequential questions and 
answers for individual scenes were generated to facilitate 3D conversations task.

\begin{figure*}
  \centering
  \includegraphics[width=1 \textwidth]{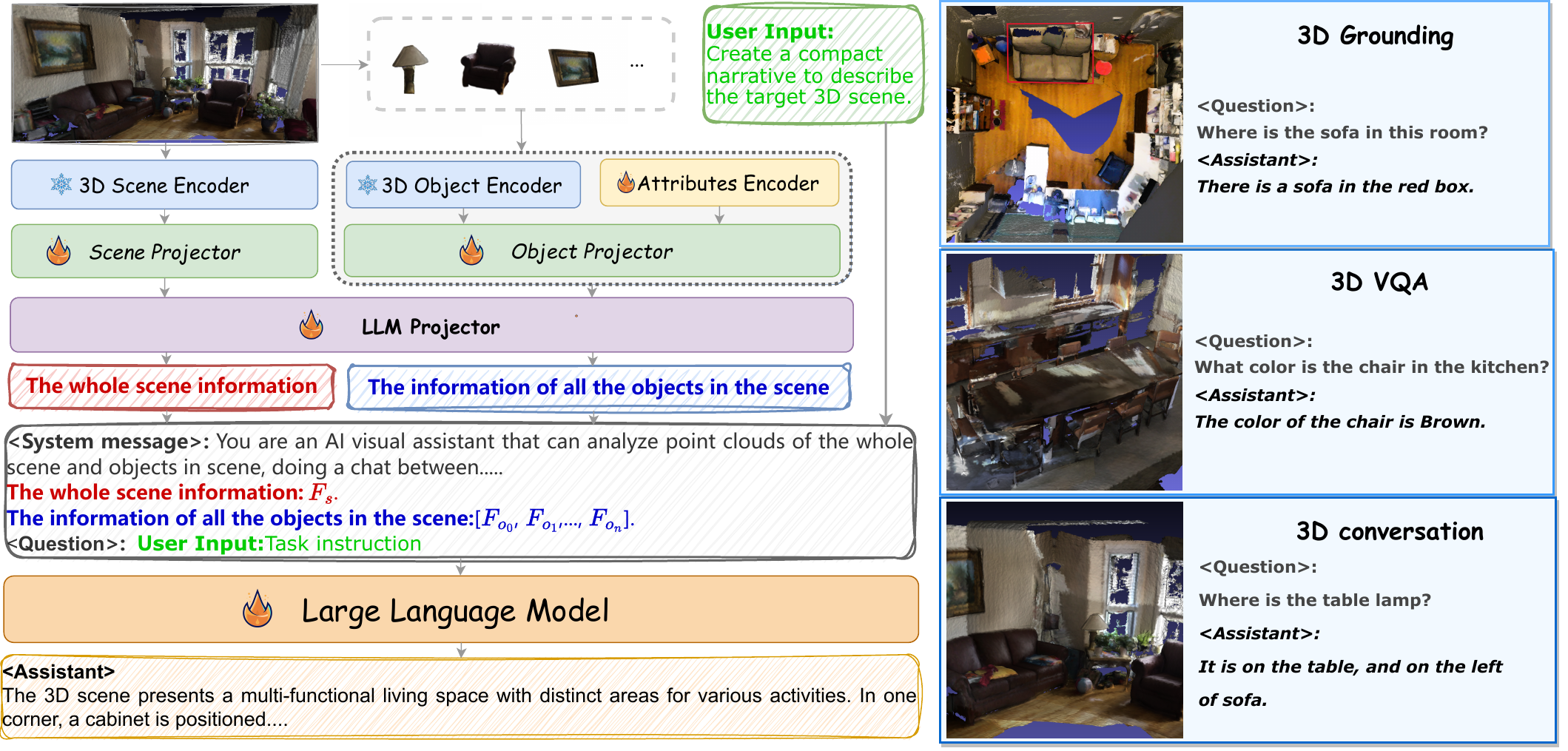}
  \caption{Overview Architecture of our method. The 3D input covers 3D scene and segmented objects. Our 3D perceiver contains frozen 3D scene encoder and 3D object encoder, trainable attributes encoder and projectors. We construct 3D vision-language prompt with system message, 3D features and task instructions. Leveraging the 3D prompt, we can fine-tune LLM efficiently by Lora to solve 3D tasks. }
  \label{fig:res}
  \vspace{-10pt}
\end{figure*}

\subsection{3D Grounding}
We construct our grounding dataset with ground truth bounding box of each object and its description in ScanRefer~\cite{chen2020scanrefer}. We use GPT to generate corresponding instructions with the prompt of ``$\ object-description$ and locate its position with the coordinate of center x, y, z and its length, width and height.", answers with ``$\ object-id$ and its bounding box". 

\section{Method}
\subsection{Overview}
Considering the time-consuming alignment stage of 3D-text in Chat-3D~\cite{wang2023chat} and 3D-LLM~\cite{hong20233d}, we explore an efficient instruction tuning method to enhance the perception and comprehension of 3D scenes by LLM. 
Without alignment stage, we propose an end-to-end fine-tuning to inject 3D information into LLM. Our overview architecture is illustrated in Figure~\ref{fig:res}.

\subsection{3D Perceiver}
The architecture comprises a 3D perceiver designed to extract essential 3D visual features for LLMs understanding the 3D scenes.
To capture both global scene information and objects detail, the 3D perceiver is consists of several key components: 
frozen 3D Scene Encoder $Es$ and 3D Object Encoder $Eo$, trainable Attributes Encoder $Ea$ and Projectors for each part. 
We use EPCL~\cite{huang2022frozen} as our 3D Scene Encoder $Es$ to extract the scene features $fs$ representing global scene information. 
For fine-grained object information, we segment the point clouds of each object in one scene and get [coordinate, size, color] as attributes of each object point cloud. 
We use the pre-trained 3D encoder~\cite{xue2023Ulip,zhou2023Uni3D} as our 3D Object Encoder $Eo$ to extract object features $fo$. 
And we follow LLaMA~\cite{touvron2023LLaMA} to use position embedding layers.We utilize some Linear Layers as our Attributes Encoder $Ea$ to encode the attributes of each object as $fa$. 
In detail, while position embedding layers encoding the coordinate as $fp$, and linear layers encoding the size $fq$ and color $fc$, we have $fa$ = concentrate[$fp$+$fq$+$fc$]. 
And we use $Ps$ to project $fs$ as $fs'$, $Po$ to project the features of each object $fo$ as $fo'$. $fs'$ and $fo'$ are normalized and have the same dimension. 
In addition, the scene features $fs'$ and the list of all the objects features [$fo'_1, fo'_2, ..., fo'_n$] are projected to the dimension that LLM can receive by LLM Projector $P_L$ respectively. And we represent them as $Fs$ and  [$Fo_1, Fo_2, ..., Fo_n$] 

\subsubsection{3D Scene Encoder}
We use EPCL~\cite{huang2022frozen} as our pre-trained 3D Scene Encoder which directly leverage the frozen CLIP\cite{radford2021learning} transformer as the encoder for point cloud tasks.
The EPCL builds a bridge between 3D modality and 2D modality by using CLIP model which is trained on billions of image-text pairs. 
We consider that EPCL can extract 3D scene features, which contains the alignment between the image and language domain.
\subsubsection{3D Object Encoder}
We use Ulip2~\cite{xue2023Ulip} and Uni3D~\cite{zhou2023Uni3D} respectively as our 3D object encoder. 
Ulip2 is a pre-trained 3D object encoder trained with 800K 3D point clouds and millions of image-text pairs.
Uni3D is an end-to-end pre-trained scalable 3D object encoder which aligns the 3D point cloud features with the image-text aligned features.
Because of our method without the alignment stage , we can get the 3D features which latently align with text  by using Ulip2 and Uni3D.

\begin{table*}[!t]
\begin{center}
\caption{Evaluation results of 3D VQA on ScanQA validation dataset.} 

\label{tab:vqa}
\begin{tabular}{c|c|c|cccccc}
\toprule
Method&Data &Alignment & EM &BLEU-1& BLEU-4 &METEOR& ROUGE &CIDEr\\
\midrule
VoteNet~\cite{ding2019votenet}+MCAN~\cite{yu2019mcan}&-& - & 17.3 &28.0& 6.2 &11.4 &29.8 &54.7 \\
ScanRefer~\cite{chen2020scanrefer}+MCAN~\cite{yu2019mcan} &-& -& 18.6&26.9 &7.9& 11.5 &30.0& 55.4 \\
\midrule
Chat-3D~\cite{wang2023chat}& -& $\checkmark$ & - &29.1 &6.4 &11.9 &28.5 &53.2  \\
3D-LLM (flamingo)& - &$\checkmark$ &20.4& 30.3& 7.2 &12.2& 32.3& 59.2  \\
3D-LLM (BLIP2-flant5) & - &$\checkmark$& 20.5 &39.3 &12.0 &14.5 &35.7 &69.4  \\
\midrule
LLaVA~\cite{liu2023LLaVA}(zero-shot) & -&$\usym{2717}$&0.0& 7.1& 0.3& 10.5& 12.3& 5.7 \\
LAMM~\cite{yin2023lamm}& 25k &$\usym{2717}$&9.82 &26.77 &5.78 &9.95 &23.64 &42.41 \\

\midrule
3DMIT(Vicuna-7b) &25k &$\usym{2717}$& 10.10 &27.86 &6.44 &10.40 &24.64 &44.38  \\
3DMIT(LLaVA1.5)+IMG &25k & $\usym{2717}$& 9.62& 26.93 &5.98 & 10.64& 24.46&  46.42\\
3DMIT(Vicuna-7b) &  75K &$\usym{2717}$& 13.04&27.63 &5.24 &10.70 &26.22 &48.03  \\

\bottomrule
\end{tabular}
\end{center}
\vspace{-10pt}
\end{table*}

\subsection{Instruction Tuning}

\subsubsection{Prompt}
After we get the 3D features contains 3D scene features $Fs$ and the list of all the objects features [$Fo_1, Fo_2, ..., Fo_n$],
we combine 3D features and text as our vision-language prompt. 

For different tasks, we utilize different system messages with 3D features. We complete the prompt with the instructions of different tasks. Here is an example and the intact and various prompts are in the supplement.
\vspace{-0.1cm}
\begin{quote}

$\#\#\#<$System message$>$: You are an AI visual assistant can analyze point clouds of the whole scene and objects in scene...$\#\#\#$ The whole scene information: $Fs$. The information of all the objects in the scene: [$Fo_1, Fo_2, ..., Fo_n$]. $<Question> $: ...
\end{quote}
\vspace{-0.1cm}

\subsubsection{Pre-trained LLM and Fine-tune}
We use Vicuna-7b~\cite{chiang2023Vicuna} and LLaVA1.5-7b~\cite{liu2023improved} as our basic LLM and MLLM respectively.
Vicuna-7b is a LLM, which is fine-tuned from LLaMA~\cite{touvron2023LLaMA}.
LLaVA1.5-7b is a MLLM which is fine-tuned with instructional vision-language data.

We use Lora~\cite{hu2021lora} to fine-tune our  basic LLM and MLLM with 3D scene-language instruction pairs.







\section{Experiments}

\subsection{Implementation Details}

We leverage pre-trained 3D scene encoder EPCL based on the ViT-L/14 of CLIP. We use pre-trained 3D object encoder Ulip2 based on the ViT-B/16 of CLIP and Uni3D based ViT-L/14 of CLIP.
For encoding object attributes, we employ a position embedding layer inspired by LLaMA for coordinating point clouds, while linear layers encode the size and color attributes of point clouds.
The architectural components of our model include linear layers for both the scene projector and LLM projector, 3D object projector comprises a normalization layer and a three-layer MLP.
To optimize the training process, we adopt the bp16 Mixed-precision Training Strategy facilitated by DeepSpeed, conducting training for 2 epochs. 
We set train batch size as 256, train micro batch size per gpu as 1, and gradient accumulation steps as 32.
We set lora\_r as 32, lora\_alpha as 32, lora\_dropout as 0.1 and learning rate as 5e-4.
The entire fine-tuning process requires approximately 7 hours on 8 NVIDIA A100 GPUs.

\subsection{Evaluation on 3D VQA}
The training datasets covers 75K scene-language instruction pairs contraining 25K 3D VQA data. We choose Vicuna-7b and LLaVA1.5-7b as our basic LLM and MLLM. We evaluate our method on ScanQA~\cite{azuma2022scanqa} validation dataset, which employs VoteNet~\cite{ding2019votenet} to generate object proposals, integrating them with text embeddings.

\subsubsection{Baselines}
The compared baselines contain 3 categories: supervised expert models, general 3D-LLMs with alignment and 3D-LLMs without alignment. 
\begin{itemize}
    \item Supervised Expert Models:  ScanRefer~\cite{chen2020scanrefer}+MCAN~\cite{yu2019mcan} and VoteNet~\cite{ding2019votenet}+MCAN\cite{yu2019mcan} leverage MCAN to detect 3D objects and incorporate them into a standard VQA model.
    \item  General 3D-LLMs with alignment: 
    3DLLM~\cite{hong20233d} algins 3D scene, 2D images with text features, and employs 2D VLMs 
    as backbone, projecting 3D features into the VLM’s input space. Chat-3D~\cite{wang2023chat} aligns 3D objects features with text but is limited to object-centric tasks.
    \item  3D-LLMs without alignment: LLaVA~\cite{liu2023LLaVA} is a MLLLM that connects a vision encoder and LLM for general purpose visual and language understanding.
    LAMM, based on Vicuna-7b, introduces the LAMM-Dataset and framework for scene understanding without the alignment stage of 3D point cloud and text.

\end{itemize}

\subsubsection{Evaluation Metrics}
We report ROUGE-L, METEOR, CIDEr for robust answer matching. We also use exact match (EM) metric, BLEU-1 and BLEU-4 metrics.
\subsubsection{Result Analysis}
As the results shown in Table~\ref{tab:vqa}, we compare our method with expert models, general 3D-LLMs with alignment and 3D-LLMs without alignment. 
Notably, when utilizing a limited training dataset of only 25k data, our method based on Vicuna-7b, surpasses 3D-LLMs without alignment, such as LAMM and LLaVA (zero shot), across all evaluated metrics. 
 As the training dataset expands to 75K instances encompassing various tasks, our method 3DMIT(Vicuna-7b) outperforms 3DMIT(Vicuna-7b) with 25K data on EM, METEOR, ROUGE and CIDEr metrics, which suggests that our method is more diverse and creative with more data. 
 The performance on BLEU-1 and BLEU-4 metrics shows a marginal decrease, indicating that our method remains competitive and robust.

Furthermore, our method achieves results comparable to expert models and 3D-LLMs with alignment, such as Chat-3D and 3D-LLM (flamingo). Notably, it outperforms VoteNet+MCAN on BLEU-4, as well as ScanRefer+MCAN and Chat-3D on BLEU-1.
We also find that the results of 3DMIT (LLaVA1.5) with multi-view image tokens closed to the performance of 3DMIT (Vicuna-7b), which proves that our method has promising transferability across different LLMs and MLLMs. 
In addition, the ablation study of introducing multi-view image tokens will be discussed in Section 5.4. We also evaluate our method on the validation dataset of multiple-choice in Supplement.

In summary, our results consistently outperform 3D-LLMs without alignment and closely approach the performance of 3D-LLMs with alignment. This substantiates the effectiveness of our proposed method in achieving competitive results across diverse evaluation metrics.

\begin{table}[!t] 
\begin{center}
\caption{Evaluation results of 3D grounding on Scanrefer validation dataset.} 

\label{tab:vg}
\scalebox{0.95}{
\begin{tabular}{c|c|cc}
\toprule
Method& Alignment &Acc@0.25 &Acc@0.5\\
\midrule

ScanRefer &-& 37.3& 24.3 \\
\midrule

3D-LLM (flamingo) &$\checkmark$ &21.2 &- \\
3D-LLM (BLIP2)  &$\checkmark$& 30.3& - \\
LLM-Grounder~\cite{yang2023llm}&  $\usym{2717}$ &17.1  & 5.3 \\
LAMM & $\usym{2717}$ & - & 3.38 \\
\midrule
3DMIT(Vicuna-7b) &$\usym{2717}$ &  10.2  &7.07 \\  
3DMIT(LLaVA1.5)  & $\usym{2717}$& 10.7&7.2\\  
\bottomrule
\end{tabular}
}
\end{center}
\vspace{-10pt}
\end{table}

\begin{table}[t]
\begin{center}
\caption{Ablation study of the scene of the multi-view image tokens.} 

\label{tab:ablation}
\begin{tabular}{c|c|cc}
\toprule
Method& Model &BLEU-1 &CIDEr\\
\midrule

w/o IMG& 3DMIT(Vicuna-7b)  &27.86   &44.38  \\
w IMG& 3DMIT(Vicuna-7b) &24.78  &40.23  \\
\midrule
w/o IMG& 3DMIT(LLaVA1.5)   &24.08   &37.45 \\
w IMG& 3DMIT(LLaVA1.5)&26.93  &46.42 \\


\bottomrule
\end{tabular}
\end{center}
\vspace{-15pt}
\end{table}

\subsection{Evaluation on 3D Visual Grounding}
In the evaluation based on the ScanRefer validation dataset, the model is given the object description. 
The model should understand the location and object description. Subsequently, it is required to provide the object identifier along with its corresponding 3D bounding box.

\subsubsection{Baselines}
\begin{itemize}
    \item ScanRefer~\cite{chen2020scanrefer} employs a pretrained VoteNet backbone integrated with a GRU to select the matching bounding box, thereby demonstrating expertise in 3D scene understanding.
    \item General 3D-LLMs with alignment: 3D-LLM~\cite{hong20233d} predicts bounding boxes as location tokens added to the LLM's vocabularies. These bounding boxes are learned from scratch, showcasing an alignment strategy to facilitate enhanced understanding.
    \item 3D-LLMs without alignment: LLM-Grounder~\cite{yang2023llm} achieves zero-shot 3D visual grounding by utilizing an open-vocabulary visual grounding tool and using GPT4 as the language assistant. LAMM~\cite{yin2023lamm} presents the LAMM-Dataset and framework for scene understanding without alignment stage of 3D point cloud and text.

\end{itemize}

\subsubsection{Valuation Metrics}
We measure 3D grounding capability by Accuracy@0.25
and @0.5, which are accuracies of bounding box predictions whose Intersection-over-Union (IoU) w.r.t. ground-truth box
exceeds 0.25 and 0.5, respectively.

\subsubsection{Result Analysis}
As the results shown in Table~\ref{tab:vg}, our comparative analysis contains our method against expert models, general 3D-LLMs with the alignment of 3D modality and text, and 3D-LLMs without alignment.

An insightful observation is that our method, based on Vicuna-7b and LLaVA1.5, outperforms 3D-LLMs without alignment, such as LLM-Grounder and LAMM in terms of Acc@0.5. 
It is proved the effectiveness of our method with answering the object identifiers and its bounding box.
And we find that LLMs face challenges in numerical and computational tasks. 
Considering expert models and 3D-LLMs with alignment outperform our method, our method has limitations in comprehending spatial location details without alignment.

However, through our experiment, we can inspire researchers to consider how to train 3D-LLMs understand location information easily and effectively.

\begin{table}[t]
\begin{center}
\caption{Ablation study of different pre-trained 3D object encoders.} 

\label{tab:3dencoder}
\begin{tabular}{c|c|cc}
\toprule
Method& Model & ROUGE &CIDEr\\
\midrule

3DMIT(Vicuna-7b) & Ulip2 & 27.63&48.03   \\
3DMIT(Vicuna-7b)&  Uni3D &25.38 &47.04 \\
\midrule
3DMIT(LLaVA1.5)&  Ulip2  &21.23   &37.45 \\
3DMIT(LLaVA1.5)&  Uni3D &22.38  &36.68 \\


\bottomrule
\end{tabular}
\end{center}
\vspace{-15pt}
\end{table}

\vspace{-5pt}
\subsection{Ablation Study}
We conduct an ablation study to discuss the effectiveness of multi-view image tokens of the scene. 
To evaluate the effectiveness of 3DMIT with the multi-view image of the scene, we train 3DMIT with multi-view tokens based on Vicuna-7b and LLaVA1.5-7b. 
The results are presented in Table~\ref{tab:ablation}. 
We find that if introducing image tokens to 3DMIT (Vicuna-7b), the performance decreases because Vicuna-7b is language model which can not understand the 3D information and 2D image simultaneously. 
By contrast, LLaVA1.5 connects the visual encoder of CLIP with LLaMA, and is fine-tuned end-to-end on instructional vision-language data. Therefore, the performance increases by introducing image tokens to 3DMIT based on LLaVA1.5-7b. We conclude that if based on MLLM such as LLaVA, introducing multi-view can improve the performance of 3DMIT.

Furthermore, we assess the effectiveness of the pre-trained 3D object encoder chosen for our model. 
 The result is shown in Table~\ref{tab:3dencoder}. 
We compare our method utilizing Ulip2 or Uni3D as our 3D object encoder, the results are closed. 
 Because both Ulip2 and Uni3D are based on CLIP for 3D-2D-text alignment, differing only in the dimensions of extracted features. 
Consequently, the pre-trained 3D object encoder used in Table~\ref{tab:vqa} and Table~\ref{tab:vg} is Ulip2 without specific declaration.

\section{Conclusion}
This paper introduces 3DMIT, an innovative and efficient 3D multi-modal instruction tuning framework designed to train LLMs in comprehending the intricacies of 3D world. Distinguishing itself from other 3D-LLMs, 3DMIT eliminates the alignment stage between 3D scenes and language, prioritizing efficiency. The framework leverages both global scene information and fine-grained object details to enhance the overall scene understanding capabilities of LLMs.
We construct a comprehensive 3D scene-language instruction dataset with GPT-API, which encompasses 75K 3D scene-language pairs for 3D tasks.
We evaluate our method on ScanQA for 3D VQA and ScanRefer for visual grounding. Our experimental results demonstrate the effectiveness of the proposed method.
We believe our insights and methodologies will inspire researchers to further explore and refine the integration of multi-modal information into LLMs directly and efficiently.


\bibliographystyle{IEEEtran}
\addtolength{\itemsep}{-10 em} 

\bibliography{icme2023template}

\begin{thebibliography}{10}
\providecommand{\url}[1]{#1}
\csname url@samestyle\endcsname
\providecommand{\newblock}{\relax}
\providecommand{\bibinfo}[2]{#2}
\providecommand{\BIBentrySTDinterwordspacing}{\spaceskip=0pt\relax}
\providecommand{\BIBentryALTinterwordstretchfactor}{4}
\providecommand{\BIBentryALTinterwordspacing}{\spaceskip=\fontdimen2\font plus
\BIBentryALTinterwordstretchfactor\fontdimen3\font minus \fontdimen4\font\relax}
\providecommand{\BIBforeignlanguage}[2]{{%
\expandafter\ifx\csname l@#1\endcsname\relax
\typeout{** WARNING: IEEEtran.bst: No hyphenation pattern has been}%
\typeout{** loaded for the language `#1'. Using the pattern for}%
\typeout{** the default language instead.}%
\else
\language=\csname l@#1\endcsname
\fi
#2}}
\providecommand{\BIBdecl}{\relax}
\BIBdecl

\bibitem{touvron2023LLaMA}
H.~Touvron, T.~Lavril, G.~Izacard, X.~Martinet, M.-A. Lachaux, T.~Lacroix, B.~Rozi{\`e}re, N.~Goyal, E.~Hambro, F.~Azhar \emph{et~al.}, ``Llama: Open and efficient foundation language models,'' \emph{arXiv preprint arXiv:2302.13971}, 2023.

\bibitem{chiang2023Vicuna}
W.-L. Chiang, Z.~Li, Z.~Lin, Y.~Sheng, Z.~Wu, H.~Zhang, L.~Zheng, S.~Zhuang, Y.~Zhuang, J.~E. Gonzalez \emph{et~al.}, ``Vicuna: An open-source chatbot impressing gpt-4 with 90\%* chatgpt quality,'' \emph{See https://vicuna. lmsys. org (accessed 14 April 2023)}, 2023.

\bibitem{zhu2023minigpt}
D.~Zhu, J.~Chen, X.~Shen, X.~Li, and M.~Elhoseiny, ``Minigpt-4: Enhancing vision-language understanding with advanced large language models,'' \emph{arXiv preprint arXiv:2304.10592}, 2023.

\bibitem{hong20233d}
Y.~Hong, H.~Zhen, P.~Chen, S.~Zheng, Y.~Du, Z.~Chen, and C.~Gan, ``3d-llm: Injecting the 3d world into large language models,'' \emph{arXiv preprint arXiv:2307.12981}, 2023.

\bibitem{yin2023lamm}
Z.~Yin, J.~Wang, J.~Cao, Z.~Shi, D.~Liu, M.~Li, L.~Sheng, L.~Bai, X.~Huang, Z.~Wang \emph{et~al.}, ``Lamm: Language-assisted multi-modal instruction-tuning dataset, framework, and benchmark,'' \emph{arXiv preprint arXiv:2306.06687}, 2023.

\bibitem{wang2023chat}
Z.~Wang, H.~Huang, Y.~Zhao, Z.~Zhang, and Z.~Zhao, ``Chat-3d: Data-efficiently tuning large language model for universal dialogue of 3d scenes,'' \emph{arXiv preprint arXiv:2308.08769}, 2023.

\bibitem{liu2023improved}
H.~Liu, C.~Li, Y.~Li, and Y.~J. Lee, ``Improved baselines with visual instruction tuning,'' \emph{arXiv preprint arXiv:2310.03744}, 2023.

\bibitem{dai2017scannet}
A.~Dai, A.~X. Chang, M.~Savva, M.~Halber, T.~Funkhouser, and M.~Nie{\ss}ner, ``Scannet: Richly-annotated 3d reconstructions of indoor scenes,'' in \emph{Proceedings of the IEEE conference on computer vision and pattern recognition}, 2017, pp. 5828--5839.

\bibitem{chen2020scanrefer}
D.~Z. Chen, A.~X. Chang, and M.~Nie{\ss}ner, ``Scanrefer: 3d object localization in rgb-d scans using natural language,'' in \emph{European conference on computer vision}.\hskip 1em plus 0.5em minus 0.4em\relax Springer, 2020, pp. 202--221.

\bibitem{huang2022frozen}
X.~Huang, S.~Li, W.~Qu, T.~He, Y.~Zuo, and W.~Ouyang, ``Frozen clip model is efficient point cloud backbone,'' \emph{arXiv preprint arXiv:2212.04098}, 2022.

\bibitem{xue2023Ulip}
L.~Xue, N.~Yu, S.~Zhang, J.~Li, R.~Mart{\'\i}n-Mart{\'\i}n, J.~Wu, C.~Xiong, R.~Xu, J.~C. Niebles, and S.~Savarese, ``Ulip-2: Towards scalable multimodal pre-training for 3d understanding,'' \emph{arXiv preprint arXiv:2305.08275}, 2023.

\bibitem{zhou2023Uni3D}
J.~Zhou, J.~Wang, B.~Ma, Y.-S. Liu, T.~Huang, and X.~Wang, ``Uni3d: Exploring unified 3d representation at scale,'' \emph{arXiv preprint arXiv:2310.06773}, 2023.

\bibitem{azuma2022scanqa}
D.~Azuma, T.~Miyanishi, S.~Kurita, and M.~Kawanabe, ``Scanqa: 3d question answering for spatial scene understanding,'' in \emph{proceedings of the IEEE/CVF conference on computer vision and pattern recognition}, 2022, pp. 19\,129--19\,139.

\bibitem{han2023imagebind}
J.~Han, R.~Zhang, W.~Shao, P.~Gao, P.~Xu, H.~Xiao, K.~Zhang, C.~Liu, S.~Wen, Z.~Guo \emph{et~al.}, ``Imagebind-llm: Multi-modality instruction tuning,'' \emph{arXiv preprint arXiv:2309.03905}, 2023.

\bibitem{radford2021learning}
A.~Radford, J.~W. Kim, C.~Hallacy, A.~Ramesh, G.~Goh, S.~Agarwal, G.~Sastry, A.~Askell, P.~Mishkin, J.~Clark \emph{et~al.}, ``Learning transferable visual models from natural language supervision,'' in \emph{International conference on machine learning}.\hskip 1em plus 0.5em minus 0.4em\relax PMLR, 2021, pp. 8748--8763.

\bibitem{ding2019votenet}
Z.~Ding, X.~Han, and M.~Niethammer, ``Votenet: A deep learning label fusion method for multi-atlas segmentation,'' in \emph{Medical Image Computing and Computer Assisted Intervention--MICCAI 2019: 22nd International Conference, Shenzhen, China, October 13--17, 2019, Proceedings, Part III 22}.\hskip 1em plus 0.5em minus 0.4em\relax Springer, 2019, pp. 202--210.

\bibitem{yu2019mcan}
Z.~Yu, J.~Yu, Y.~Cui, D.~Tao, and Q.~Tian, ``Deep modular co-attention networks for visual question answering,'' in \emph{Proceedings of the IEEE/CVF conference on computer vision and pattern recognition}, 2019, pp. 6281--6290.

\bibitem{liu2023LLaVA}
H.~Liu, C.~Li, Q.~Wu, and Y.~J. Lee, ``Visual instruction tuning,'' \emph{arXiv preprint arXiv:2304.08485}, 2023.

\bibitem{hu2021lora}
E.~J. Hu, Y.~Shen, P.~Wallis, Z.~Allen-Zhu, Y.~Li, S.~Wang, L.~Wang, and W.~Chen, ``Lora: Low-rank adaptation of large language models,'' \emph{arXiv preprint arXiv:2106.09685}, 2021.

\bibitem{yang2023llm}
J.~Yang, X.~Chen, S.~Qian, N.~Madaan, M.~Iyengar, D.~F. Fouhey, and J.~Chai, ``Llm-grounder: Open-vocabulary 3d visual grounding with large language model as an agent,'' \emph{arXiv preprint arXiv:2309.12311}, 2023.

\end{thebibliography}

\newpage 

\appendix 


\textbf{{\fontsize{16pt}{14pt}\selectfont
Supplemental Material}}

\vspace{0.5cm}
In this supplemental material, we report more experiments about our 3DMIT, including case study of our method, results of more experiments and more visual prompt examples.

\section{Case Study of our method}
\begin{figure}[h]
  \centering
  \includegraphics[width=0.5 \textwidth]{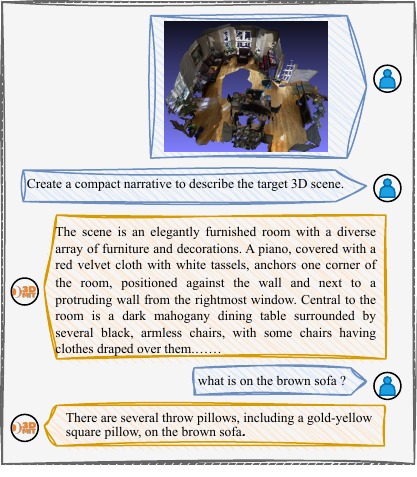}
  \caption{The case study of 3D conversation and 3D caption task.}
  \label{fig:case1}
\end{figure}

\begin{figure}[h]
  \centering
  \includegraphics[width=0.5 \textwidth]{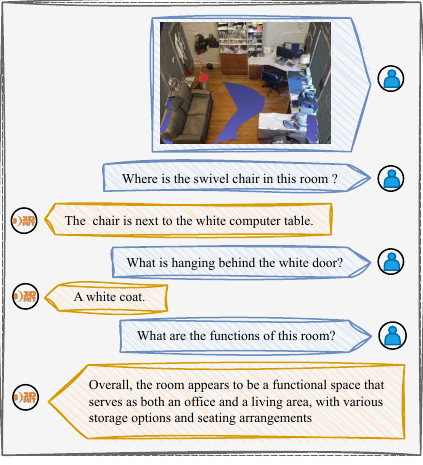}
  \caption{The case study of 3D conversation and 3D VQA task.}
  \label{fig:case2}
\end{figure}

Figure~\ref{fig:case1} shows the case study of 3D conversation and 3D caption task. Figure~\ref{fig:case2} shows the case study of 3D conversation and 3D VQA task.

\section{Evaluation results of 3D Multiple-choice}
We construct the training dataset and validation dataset based on ScanQA dataset, which employs VoteNet to generate object proposals, integrating them with text embeddings. We use GPT to generate choices for multiple-choice question answering problem on 3D scene dataset.
The result is presented in Table~\ref{tab:choice}. 
We utilize the accuracy of multiple-choice to evaluate our results, which is represented as Acc.
We can find that the Acc of all of our method is over 90\%. It can prove that the effectiveness of our method in 3D VQA and its corresponding multiple-choice format Question-Answer task. Furthermore, with the results and conclusions in the paper, we can prove our method 3DMIT provide a promising ability of understanding of the 3D world.

\begin{table}[h]
\begin{center}
\caption{Evaluation results of 3D Multiple-choice on ScanQA validation dataset} \label{tab:choice}
\begin{tabular}{c|c|c|c}
\hline
Method& Alignment &Object Encoder &Acc\\
\hline
3DMIT(Vicuna-7b)& $\usym{2717}$& Ulip2 & 91.17\\
3DMIT(Vicuna-7b)& $\usym{2717}$& Uni3D & 90.68\\
3DMIT(LLaVA1.5)  & $\usym{2717}$& Uni3D & 93.53\\

\hline
\end{tabular}
\end{center}
\end{table}

\section{visual prompt examples}
\begin{figure}[h]
  \centering
  \includegraphics[width=0.44 \textwidth]{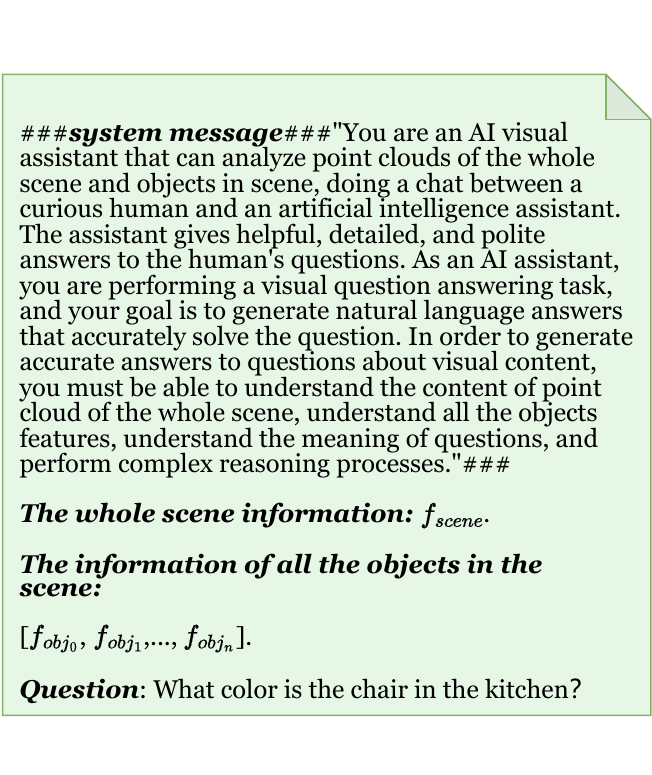}
  \caption{The 3D vision-language prompt for VQA with system message, 3D features and task instructions. }
  \label{fig:vqa}
\end{figure}

\begin{figure}[h]
  \centering
  \includegraphics[width=0.44 \textwidth]{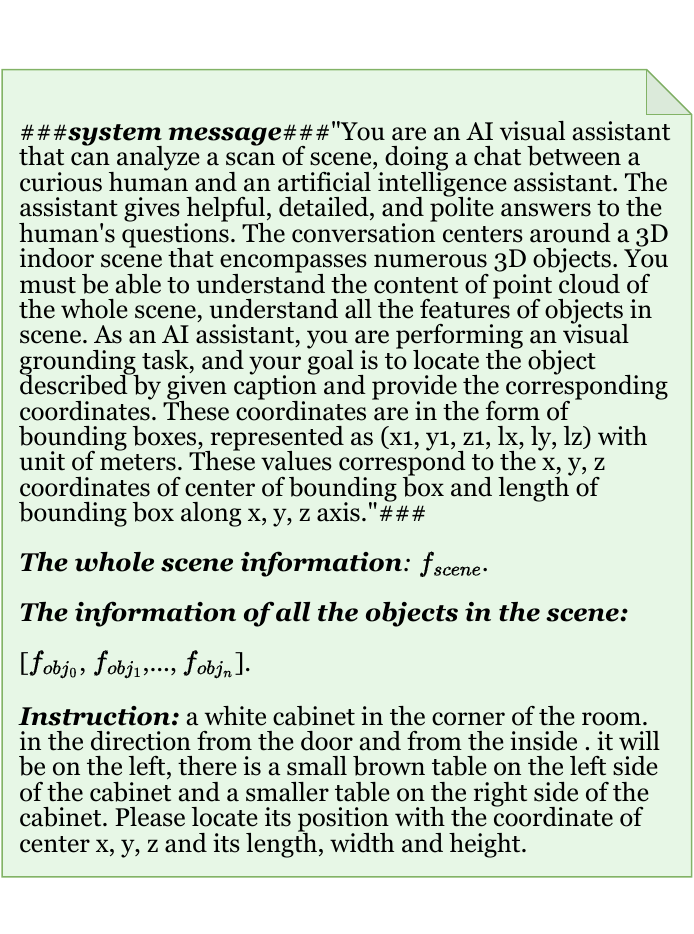}
  \caption{The 3D vision-language prompt for Visual grounding with system message, 3D features and task instructions.}
  \label{fig:vg}
\end{figure}

 For different 3D tasks, we construct different 3D prompt with various system message and task instructions. Figure~\ref{fig:vqa} shows the detailed 3D vision-language prompt for VQA task, and Figure~\ref{fig:vg} shows the detailed 3D vision-language prompt for 3D visual grounding task.

\end{document}